\documentclass{article}

    \PassOptionsToPackage{numbers, compress}{natbib}

    \usepackage[final]{neurips_2021}

\usepackage[utf8]{inputenc} 
\usepackage[T1]{fontenc}    
\usepackage{hyperref}       
\usepackage{url}            
\usepackage{booktabs}       
\usepackage{amsfonts}       
\usepackage{nicefrac}       
\usepackage{microtype}      
\usepackage{xcolor}         
\usepackage[pdftex]{graphicx}
\usepackage{amsmath}
\usepackage{amssymb}
\usepackage{wrapfig}
\usepackage{booktabs}

\DeclareMathOperator*{\E}{\mathbb{E}}
\def\4#1{\mathcal{#1}}
\def\2#1{\mathbf{#1}}
\def\3#1{\texttt{#1}}
\def\5#1{\textit{#1}}
\def\kl#1#2{\text{KL}\big(#1\|#2\big)}

\DeclareMathOperator*{\argmax}{\text{argmax}}

\newif\ifblind
\blindfalse

\title{CogView: Mastering Text-to-Image Generation via Transformers}

\author{%
  Ming Ding\textsuperscript{\textdagger}, Zhuoyi Yang\textsuperscript{\textdagger}, Wenyi Hong\textsuperscript{\textdagger}, Wendi Zheng\textsuperscript{\textdagger}, Chang Zhou\textsuperscript{$\ddagger$}, Da Yin\textsuperscript{\textdagger},\\ \textbf{Junyang Lin\textsuperscript{$\ddagger$}, Xu Zou\textsuperscript{\textdagger}, Zhou Shao\textsuperscript{$\spadesuit$}, Hongxia Yang\textsuperscript{$\ddagger$}, Jie Tang\textsuperscript{\textdagger}}\textsuperscript{$\spadesuit$} \\
  \textsuperscript{\textdagger}Tsinghua University \  
  \textsuperscript{$\ddagger$}DAMO Academy, Alibaba Group \ \textsuperscript{$\spadesuit$}BAAI \\
  \texttt{\{dm18@mails, jietang@mail\}.tsinghua.edu.cn}\\
}

\begin{document}

\maketitle 

\begin{abstract}
  Text-to-Image generation in the general domain has long been an open problem, which requires both a powerful generative model and cross-modal understanding. We propose CogView, a 4-billion-parameter Transformer with VQ-VAE tokenizer to advance this problem. We also demonstrate the finetuning strategies for various downstream tasks, e.g. style learning, super-resolution, text-image ranking and fashion design, and methods to stabilize pretraining, e.g. eliminating NaN losses. CogView achieves the state-of-the-art FID on the blurred MS COCO dataset, outperforming previous GAN-based models and a recent similar work DALL-E.
  \ifblind
\else
  \footnote{Codes and models are at \url{https://github.com/THUDM/CogView}. We also have a demo website of our latest model at \url{https://wudao.aminer.cn/CogView/index.html} (without post-selection).}
  \fi
\end{abstract}
\begin{figure*}[h]  
  \vspace*{-2mm}
  \includegraphics[width=\textwidth]{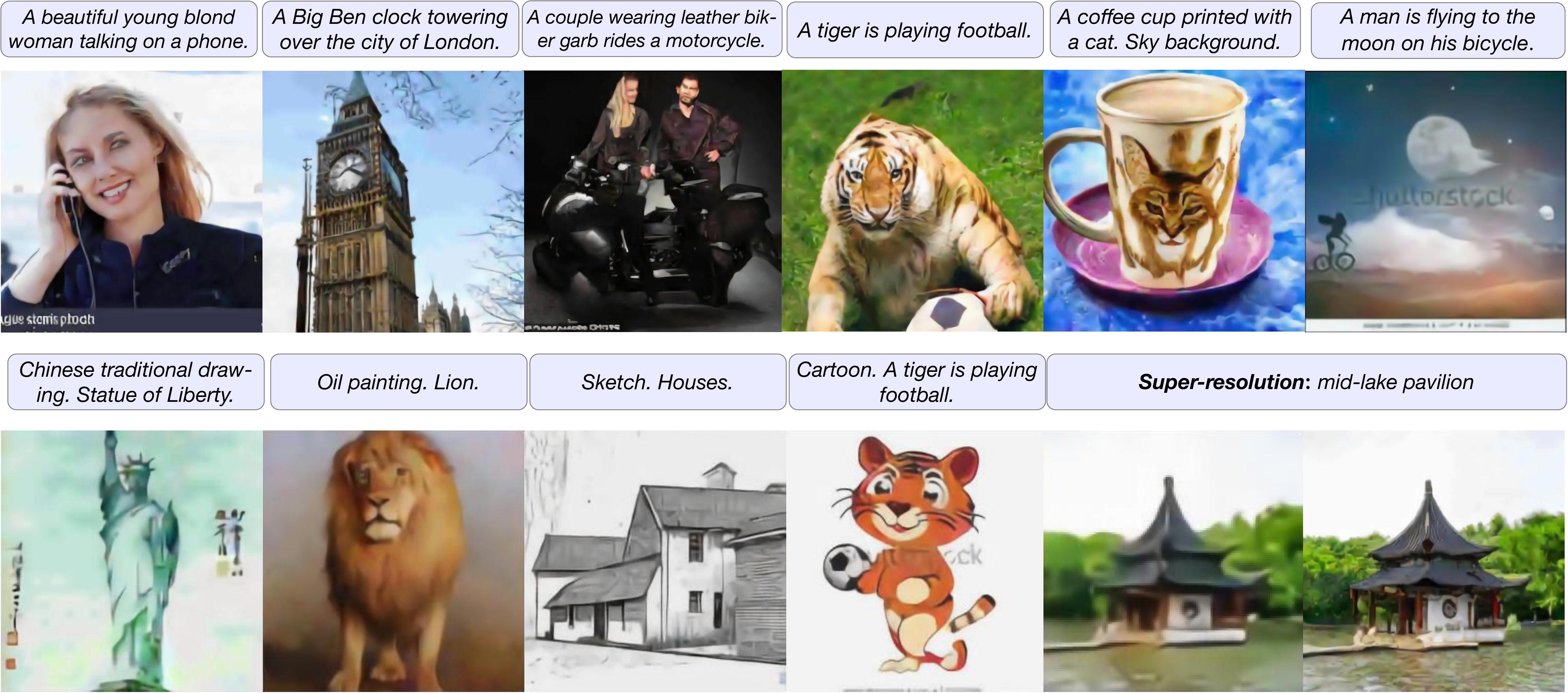}\label{fig:teaser}
  \caption{Samples generated by CogView. The text in the first line is either from MS COCO (outside our training set) or user queries on our demo website. The images in the second line are finetuned results for different styles or super-resolution. The actual input text is in Chinese, which is translated into English here for better understanding. More samples for captions from MS COCO are included in \ifblind\else Appendix~\ref{app:case}.\fi}
\end{figure*}
\vspace*{-2mm}

\section{Introduction}\label{sec:intro}
\begin{quote}
\emph{
``There are two things for a painter, the eye and the mind... 
eyes, through which we view the nature; brain, in which we organize sensations by logic for meaningful expression.'' (Paul C\'ezanne~\cite{cezanne})}
\end{quote}
As contrastive self-supervised pretraining has revolutionized computer vision (CV)~\cite{he2020momentum,grill2020bootstrap,chen2020simple,liu2020self}, visual-language pretraining, which brings high-level semantics to images, is becoming the next frontier of visual understanding~\cite{radford2021learning,lin2021m6, ramesh2021zero}. Among various pretext tasks, text-to-image generation expects the model to (1) disentangle shape, color, gesture and other features from pixels, (2) understand the input text, (2) align objects and features with corresponding words and their synonyms and (4) learn complex distributions to generate the overlapping and composite of different objects and features, which, like painting, is beyond basic visual functions (related to eyes and the V1--V4 in brain~\cite{grill2004human}), requiring a higher-level cognitive ability (more related to the angular gyrus in brain~\cite{bonnici2016multimodal}).

The attempts to teach machines text-to-image generation can be traced to the early times of deep generative models, when \citet{mansimov2015generating} added text information to DRAW~\cite{gregor2015draw}. Then Generative Adversarial Nets~\cite{goodfellow2014generative} (GANs) began to dominate this task. \citet{reed2016generative} fed the text embeddings to both generator and discriminator as extra inputs. StackGAN~\cite{zhang2017stackgan} decomposed the generation into a sketch-refinement process. AttnGAN~\cite{xu2018attngan} used attention on words to focus on the corresponding subregion. ObjectGAN~\cite{li2019object} generated images following a text$\rightarrow$boxes$\rightarrow$layouts$\rightarrow$image process. DM-GAN~\cite{zhu2019dm} and DF-GAN~\cite{tao2020df} introduced new architectures, e.g. dyanmic memory or deep fusion block, for better image refinement. Although these GAN-based models can perform reasonable synthesis in simple and domain-specific dataset, e.g. Caltech-UCSD Birds 200 (CUB), the results on complex and domain-general scenes, e.g. MS COCO~\cite{lin2014microsoft}, are far from satisfactory.

Recent years have seen a rise of the auto-regressive generative models. Generative Pre-Training (GPT) models~\cite{radford2019language,brown2020language} leveraged Transformers~\cite{vaswani2017attention} to learn language models in large-scale corpus, greatly promoting the performance of natural language generation and few-shot language understanding~\cite{liu2021gpt}. Auto-regressive model is not nascent in CV. PixelCNN, PixelRNN~\cite{van2016pixel} and Image Transformer~\cite{parmar2018image} factorized the probability density function on an image over its sub-pixels (color channels in a pixel) with different network backbones, showing promising results. However, a real image usually comprises millions of sub-pixels, indicating an unaffordable amount of computation for large models. Even the biggest pixel-level auto-regressive model, ImageGPT~\cite{chen2020generative}, was pretrained on ImageNet at a max resolution of only $96\times 96$.  

The framework of Vector Quantized Variational AutoEncoders (VQ-VAE)~\cite{van2017neural} alleviates this problem. VQ-VAE trains an encoder to compress the image into a low-dimensional discrete latent space, and a decoder to recover the image from the hidden variable in the stage 1. Then in the stage 2, an auto-regressive model (such as PixelCNN~\cite{van2016pixel}) learns to fit the prior of hidden variables. This discrete compression loses less fidelity than direct downsampling, meanwhile maintains the spatial relevance of pixels. Therefore, VQ-VAE revitalized the auto-regressive models in CV~\cite{razavi2019generating}. Following this framework, \citet{esser2020taming} used Transformer to fit the prior and further switches from $L_2$ loss to GAN loss for the decoder training, greatly improving the performance of domain-specific unconditional generation.

The idea of CogView comes naturally: large-scale generative joint pretraining for both text and image (from VQ-VAE) tokens. We collect 30 million high-quality (Chinese) text-image pairs and pretrain a Transformer with 4 billion parameters. However, large-scale text-to-image generative pretraining could be very unstable due to the heterogeneity of data. We systematically analyze the reasons and solved this problem by the proposed \emph{Precision Bottleneck Relaxation} and \emph{Sandwich Layernorm}. As a result, CogView greatly advances the quality of text-to-image generation. 

A recent work DALL-E~\cite{ramesh2021zero} independently proposed the same idea, and was released earlier than CogView. Compared with DALL-E, CogView steps forward on the following four aspects:
\begin{itemize}
    \item CogView outperforms DALL-E and previous GAN-based methods at a large margin according to the Fr\'echet Inception Distance (FID)~\cite{heusel2017gans} on blurred MS COCO, and is the first open-source large text-to-image transformer. 
    \item Beyond zero-shot generation, we further investigate the potential of finetuning the pretrained CogView. CogView can be adapted for diverse downstream tasks, such as style learning (domain-specific text-to-image), super-resolution (image-to-image), image captioning (image-to-text), and even text-image reranking.
    \item The finetuned CogView enables self-reranking for post-selection, and gets rid of an additional CLIP model~\cite{radford2021learning} in DALL-E. It also provides a new metric \emph{Caption Loss} to measure the quality and accuracy for text-image generation at a finer granularity than FID and Inception Score (IS)~\cite{salimans2016improved}.  
    \item We proposed PB-relaxation and Sandwich-LN to stabilize the training of large Transformers on complex datasets. These techniques are very simple and can eliminate overflow in forwarding (characterized as NaN losses), and make CogView able to be trained with \emph{almost FP16} (O2\footnote{meaning that all computation, including forwarding and backwarding are in FP16 without any conversion, but the optimizer states and the master weights are FP32.}). They can also be generalized to the training of other transformers.
\end{itemize}
\section{Method}

\subsection{Theory}
In this section, we will derive the theory of CogView from VAE\footnote{In this paper, \textbf{bold} font denotes a random variable, and regular font denotes a concrete value.
\ifblind
\else
See this comprehensive tutorial~\cite{dingroad} for the basics of VAE.
\fi
 }~\cite{kingma2013auto}: \emph{CogView optimizes the Evidence Lower BOund (ELBO) of joint likelihood of image and text.} The following derivation will turn into a clear re-interpretation of VQ-VAE if without text $\2t$.

Suppose the dataset $(\2{X},\2T)=\{x_i, t_i\}_{i=1}^N$ consists of $N$ i.i.d. samples of image variable $\2x$ and its description text variable $\2t$. We assume the image $\2x$ can be generated by a random process involving a latent variable $\2z$: (1) $t_i$ is first generated from a prior $p(\2t;\theta)$. (2) $z_i$ is then generated from the conditional distribution $p(\2z|\2t=t_i;\theta)$. (3) $x_i$ is finally generated from $p(\2x|\2z=z_i;\psi)$. We will use a shorthand form like $p(x_i)$ to refer to $p(\2x=x_i)$ in the following part.

Let $q(\2z|x_i;\phi)$ be the variational distribution, which is the output of the encoder $\phi$ of VAE.
The log-likelihood and the evidence lower bound (ELBO) can be written as:
\begin{align}
  &\log p(\2{X},\2T;\theta,\psi)=\sum_{i=1}^N\log p(t_i;\theta) + \sum_{i=1}^N\log p(x_i|t_i;\theta,\psi)\\
  &\geq-\sum_{i=1}^N\bigg(\underbrace{-\log p(t_i;\theta)}_{\text{NLL loss for text}}
  +\underbrace{\E_{z_i\sim q(\2z|x_i;\phi)}[-\log p(x_i|z_i;\psi)]}_{\text{reconstruction loss}} + \underbrace{\kl{q(\2z|x_i;\phi)}{p(\2z|t_i;\theta)}}_{\text{KL between $q$ and (text conditional) prior}}\bigg).\label{eq:elbo}
\end{align}
The framework of VQ-VAE differs with traditional VAE mainly in the KL term. 
Traditional VAE fixes the prior $p(\2z|t_i; \theta)$, usually as $\4{N}(0,\2I)$, and learns the encoder $\phi$. However, it leads to \emph{posterior collapse}~\cite{he2018lagging}, meaning that $q(\2z|x_i;\phi)$ sometimes collapses towards the prior. VQ-VAE turns to fix $\phi$ and fit the prior $p(\2z|t_i; \theta)$ with another model parameterized by $\theta$. This technique eliminates posterior collapse, because the encoder $\phi$ is now only updated for the optimization of the reconstruction loss. In exchange, the approximated posterior $q(\2z|x_i;\phi)$ could be very different for different $x_i$, so we need a very powerful model for $p(\2z|t_i; \theta)$ to minimize the KL term.

Currently, the most powerful generative model, Transformer (GPT), copes with sequences of tokens over a discrete codebook. To use it, we make $\2z \in \{0,...,|V|-1\}^{h\times w}$, where $|V|$ is the size of codebook and $h\times w$ is the number of dimensions of $\2z$. The sequences $z_i$ can be either sampled from $q(\2z|x_i;\phi)$, or directly $z_i=\argmax_\2z q(\2z|x_i;\phi)$. We choose the latter for simplicity, so that $q(\2z|x_i;\phi)$ becomes a one-point distribution on $z_i$. The Equation~\eqref{eq:elbo} can be rewritten as:
\begin{align}
  -\sum_{i=1}^N\bigg(\underbrace{\E_{z_i\sim q(\2z|x_i;\phi)}[-\log p(x_i|z_i;\psi)]}_{\text{reconstruction loss}} \underbrace{-\log p(t_i;\theta)}_{\text{NLL loss for text}}
  \underbrace{-\log p(z_i| t_i; \theta)}_{\text{NLL loss for $\2z$}}\bigg).\label{eq:vqvae}
\end{align}
The learning process is then divided into two stages: (1) The encoder $\phi$ and decoder $\psi$ learn to minimize the reconstruction loss. (2) A single GPT optimizes the two negative log-likelihood (NLL) losses by concatenating text $t_i$ and $z_i$ as an input sequence.  

As a result, the first stage degenerates into a pure discrete Auto-Encoder, serving as an \emph{image tokenizer} to transform an image to a sequence of tokens; the GPT in the second stage undertakes most of the modeling task. Figure~\ref{fig:frame} illustrates the framework of CogView.\label{sec:theory}

\subsection{Tokenization}\label{sec:tokenizer}
In this section, we will introduce the details about the tokenizers in CogView and a comparison about different training strategies about the image tokenizer (VQVAE stage 1). 

Tokenization for text is already well-studied, e.g. BPE~\cite{gage1994new} and SentencePiece~\cite{kudo-richardson-2018-sentencepiece}. In CogView,  we ran SentencePiece on a large Chinese corpus to extract 50,000 text tokens. 

The image tokenizer is a discrete Auto-Encoder, which is similar to the stage 1 of VQ-VAE~\cite{van2017neural} or d-VAE~\cite{ramesh2021zero}. More specifically, the Encoder $\phi$ maps an image $x$ of shape $H\times W\times 3$ into $\text{Enc}_{\phi}(x)$  of shape $ h \times w \times d$, and then each $d-$dimensional vector is quantized to a \emph{nearby} embedding in a learnable codebook $\{v_0, ..., v_{|V|-1}\}, \forall v_k \in \mathbb{R}^{d}$. The quantized result can be represented by $h\times w$ indices of embeddings, and then we get the latent variable $\2z \in \{0,...,|V|-1\}^{h\times w}$. The Decoder $\psi$ maps the quantized vectors back to a (blurred) image to reconstruct the input. In our 4B-parameter CogView, $|V|=8192, d=256, H=W=256, h=w=32$. 

The training of the image tokenizer is non-trivial due to the existence of discrete selection. Here we introduce four methods to train an image tokenizer.

\begin{itemize}
  \item \emph{The nearest-neighbor mapping, straight-through estimator}~\cite{bengio2013estimating}, which is proposed by the original VQVAE. A common concern of this method~\cite{ramesh2021zero} is that, when the codebook is large and not initialized carefully, only a few of embeddings will be used due to the curse of dimensionality. We did not observe this phenomenon in the experiments.
\item \emph{Gumbel sampling, straight-through estimator}. If we follow the original VAE to reparameterize a categorical distribution of latent variable $\2z$ based on distance between vectors, i.e. $p(\2z_{i\times w + j}=v_k | x) = \frac{e^{-\|v_k-\text{Enc}_{\phi}(x)_{ij}\|_2/\tau}}{\sum_{k=0}^{|V|-1}e^{-\|v_k-\text{Enc}_{\phi}(x)_{ij}\|_2/\tau}}$, an unbiased sampling strategy is $z_{i\times w+j}=\argmax_k g_k -\|v_k-\text{Enc}_{\phi}(x)_{ij}\|_2/\tau,\  g_k \sim \text{Gumbel}(0,1),$
where the temperature $\tau$ is gradually decreased to 0. 
We can further use the differentiable softmax to approximate the one-hot distribution from argmax. DALL-E adopts this method with many other tricks to stabilize the training.
\item \emph{The nearest-neighbor mapping, moving average}, where each embedding in the codebook is updated periodically during training as the mean of the vectors recently mapped to it~\cite{van2017neural}.
\item \emph{The nearest-neighbor mapping, fixed codebook}, where the codebook is fixed after initialized.
\end{itemize}

\begin{wrapfigure}[13]{r}{0.42\textwidth}
  \centering
  \vspace{-5mm}%
  \includegraphics[width=0.42\textwidth]{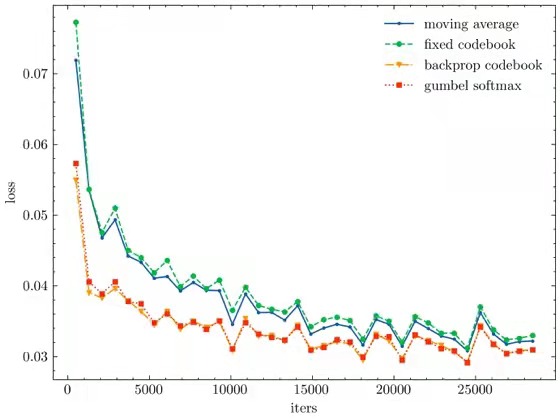} 
  \caption{$L_2$ loss curves during training image tokenizers. All the above methods finally converge to a similar loss level. }
  \label{fig:vqvae}
\end{wrapfigure}
\textbf{Comparison.} To compare the methods, we train four image tokenizers with the same architecture on the same dataset and random seed, and demonstrate the loss curves in Figure~\ref{fig:vqvae}. 
We find that all the methods are basically evenly matched, meaning that the learning of the embeddings in the codebook is not very important, if initialized properly. In pretraining, we use the tokenizer of moving average method.

The introduction of \textbf{data} and more details about tokenization are in 
\ifblind
\else
Appendix~\ref{app:data}.
\fi

\subsection{Auto-regressive Transformer}
\begin{figure*}[t] 
  \includegraphics[width=\textwidth]{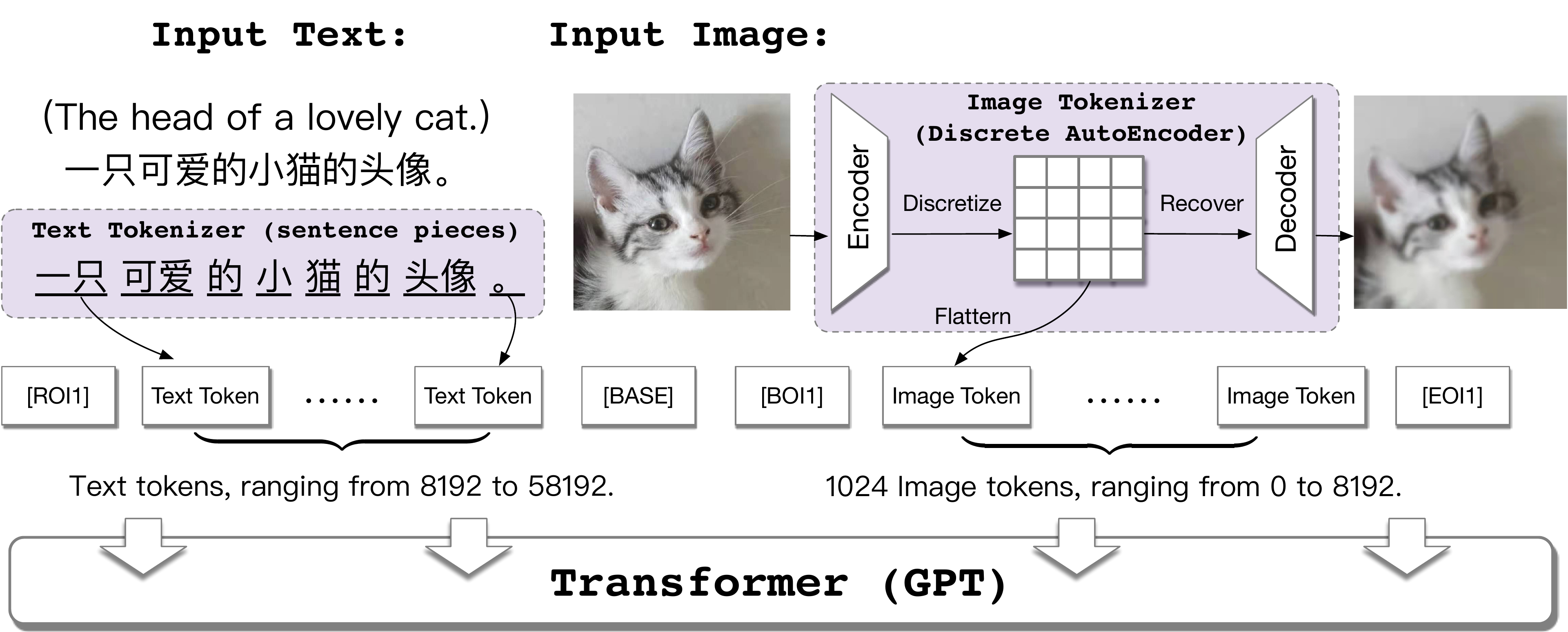}
  \caption{The framework of CogView. \texttt{[ROI1]}, \texttt{[BASE1]}, etc., are seperator tokens.}\label{fig:frame}
  \vspace{-4mm}%
\end{figure*}
The backbone of CogView is a unidirectional Transformer (GPT). The Transformer has 48 layers, with the hidden size of 2560, 40 attention heads and 4 billion parameters in total. As shown in Figure~\ref{fig:frame}, four seperator tokens, \texttt{[ROI1]} (reference text of image), \texttt{[BASE]}, \texttt{[BOI1]} (beginning of image), \texttt{[EOI1]} (end of image) are added to each sequence to indicate the boundaries of text and image. All the sequences are clipped or padded to a length of 1088.

The pretext task of pretraining is left-to-right token prediction, a.k.a. language modeling. Both image and text tokens are equally treated. DALL-E~\cite{ramesh2021zero} suggests to lower the loss weight of text tokens; on the contrary, during small-scale experiments we surprisingly find the text modeling is the key for the success of text-to-image pretraining. If the loss weight of text tokens is set to zero, the model will fail to find the connections between text and image and generate images totally unrelated to the input text. We hypothesize that text modeling abstracts knowledge in hidden layers, which can be efficiently exploited during the later image modeling. 

We train the model with batch size of 6,144 sequences (6.7 million tokens per batch) for 144,000 steps on 512 V100 GPUs (32GB). The parameters are updated by Adam with max 
$lr=3\times 10^{-4}, \beta_1=0.9, \beta_2=0.95, \text{weight decay}=4\times 10^{-2}$. The learning rate warms up during the first 2\% steps and decays with cosine annealing~\cite{loshchilov2016sgdr}. With hyperparameters in an appropriate range, we find that the training loss mainly depends on the total number of trained tokens (tokens per batch $\times$ steps), which means that doubling the batch size (and learning rate) results in a very similar loss if the same number of tokens are trained. Thus, we use a relatively large batch size to improve the parallelism and reduce the percentage of time for communication. We also design a three-region sparse attention to speed up training and save memory without hurting the performance, which is introduced in 
\ifblind\else Appendix~\ref{app:sparse}.\fi

\subsection{Stabilization of training}\label{sec:stable}
Currently, pretraining large models (>2B parameters) usually relies on 16-bit precision to save GPU memory and speed up the computation. Many frameworks, e.g. DeepSpeed ZeRO~\cite{rasley2020deepspeed}, even only support FP16 parameters. However, text-to-image pretraining is very unstable under 16-bit precision. Training a 4B ordinary pre-LN Transformer will quickly result in NaN loss within 1,000 iterations. To stabilize the training is the most challenging part of CogView, which is well-aligned with DALL-E.  

We summarize the solution of DALL-E as to \emph{tolerate} the numerical problem of training. Since the values and gradients vary dramatically in scale in different layers, they propose a new mixed-precision framework \emph{per-resblock loss scaling} and store all gains, biases, embeddings, and unembeddings in 32-bit precision, with 32-bit gradients. This solution is complex, consuming extra time and memory and not supported by most current training frameworks. 

CogView 
instead \emph{regularizes} the values. We find that there are two kinds of instability: overflow (characterized by NaN losses) and underflow (characterized by diverging loss). The following techniques are proposed to solve them. 

\textbf{Precision Bottleneck Relaxation (PB-Relax).}
After analyzing the dynamics of training, we find that overflow always happens at two \emph{bottleneck} operations, the final LayerNorm or attention. 
\begin{itemize}
  \item In the deep layers, the values of the outputs could \emph{explode} to be as large as $10^4\sim10^5$, making the variation in LayerNorm overflow. Luckily, as $\text{LayerNorm}(x) = \text{LayerNorm}(x / \max(x))$, we can relax this bottleneck by dividing the maximum first\footnote{We cannot directly divide $x$ by a large constant, which will lead to underflow in the early stage of training.}.
  \item The attention scores $Q^TK/\sqrt{d}$ could be significantly larger than input elements, and result in overflow. Changing the computational order into $Q^T(K/\sqrt{d})$ alleviates the problem. To eliminate the overflow, we notice that $\text{softmax}(Q^TK/\sqrt{d}) = \text{softmax}(Q^TK/\sqrt{d} - \text{constant})$, meaning that we can change the computation of attention into 
\begin{align}
  \text{softmax}(\frac{Q^TK}{\sqrt{d}})
  =\text{softmax}\bigg(\big(\frac{Q^T}{\alpha\sqrt{d}}K - \max(\frac{Q^T}{\alpha\sqrt{d}}K)\big)\times \alpha\bigg),
\end{align}
where $\alpha$ is a big number, e.g. 
$\alpha=32$.\footnote{The max must be at least head-wise, because the values vary greatly in different heads.} In this way, the maximum (absolute value) of attention scores are also divided by $\alpha$ to prevent it from overflow. A detailed analysis about the attention in CogView is in \ifblind\else Appendix~\ref{app:attn}.\fi
\end{itemize}
\begin{figure*}[t] 
  \includegraphics[width=\textwidth]{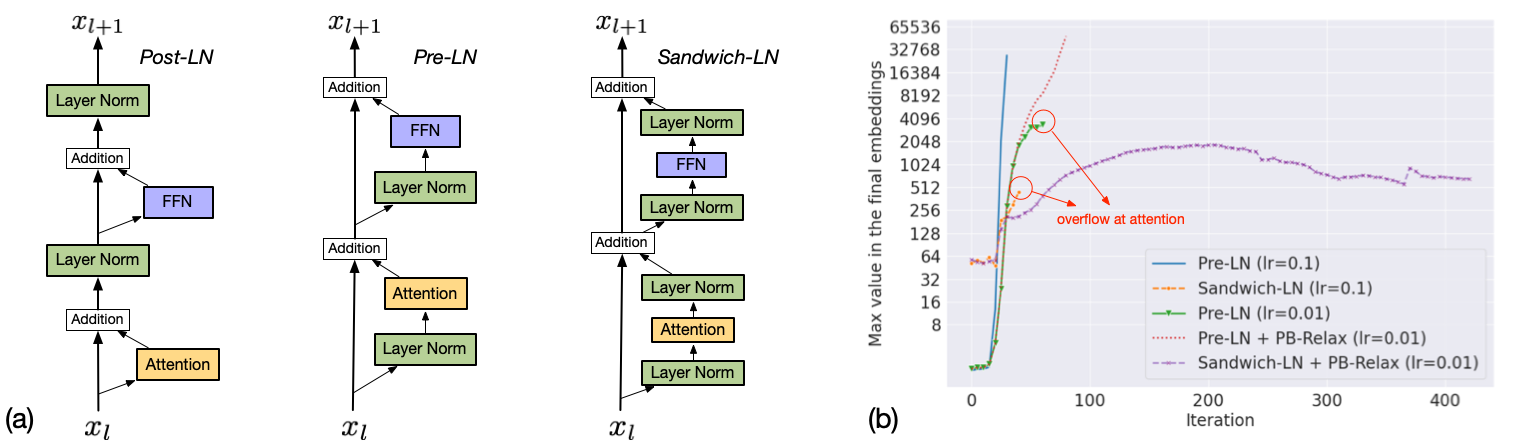}
  \caption{(a) Illustration of different LayerNorm structures in Transformers. Post-LN is from the original paper; Pre-LN is the most popular structure currently; Sandwich-LN is our proposed structure to stabilize training. (b) The numerical scales in our toy experiments with 64 layers and a large learning rate. Trainings without Sandwich-LN overflow in main branch; trainings without PB-relax overflow in attention; Only the training with both can continue. }\label{fig:stable}
  \vspace{-4mm}%

\end{figure*}
\textbf{Sandwich LayerNorm (Sandwich-LN). }The LayerNorms~\cite{ba2016layer} in Transformers are essential for stable training. Pre-LN~\cite{xiong2020layer} is proven to converge faster and more stable than the original Post-LN, and becomes the default structure of Transformer layers in recent works. However, it is not enough for text-to-image pretraining. The output of LayerNorm $\frac{(x - \bar x)\sqrt{d}}{\sqrt{\sum_i (x_i - \bar x)^2}}\gamma + \beta$ is basically proportional to the square root of the hidden size of $x$, which is $\sqrt{d} = \sqrt{2560} \approx 50$ in CogView. If input values in some dimensions are obviously larger than the others -- which is true for Transformers -- output values in these dimensions will also be large ($10^1\sim 10^2$). In the residual branch, these large values are magnified and be added back to the main branch, which aggravates this phenomenon in the next layer, and finally causes the \emph{value explosion} in the deep layers. 

This reason behind value explosion inspires us to restrict the layer-by-layer aggravation. We propose Sandwich LayerNorm, which also adds a LayerNorm at the end of each residual branch. Sandwich-LN ensures the scale of input values in each layer within a reasonable range, and experiments on training 500M model shows that its influence on convergence is negligible. Figure~\ref{fig:stable}(a) illustrates different LayerNorm structures in Transformers. 

\textbf{Toy Experiments.} Figure~\ref{fig:stable}(b) shows the effectiveness of PB-relax and Sandwich-LN with a toy experimental setting, since training many large models for verification is not realistic. We find that \emph{deep transformers} (64 layers, 1024 hidden size), \emph{large learning rates} (0.1 or 0.01), \emph{small batch size} (4) can simulate the value explosion in training with reasonable hyperparameters. PB-relax + Sandwich-LN can even stabilize the toy experiments.

\textbf{Shrink embedding gradient.} Although we did not observe any sign of underflow after using Sandwich-LN, we find that the gradient of token embeddings is much larger than that of the other parameters, so that simply shrinking its scale by $\alpha=0.1$ increases the dynamic loss scale to further prevent underflow, which can be implemented by \texttt{emb=emb*alpha+emb.detach()*(1-alpha)} in Pytorch. It seems to slow down the updating of token embeddings, but actually does not hurt performance in our experiments, which also corresponds to a recent work MoCo v3~\cite{chen2021empirical}. 

\textbf{Discussion. }The PB-relax and Sandwich-LN successfully stabilize the training of CogView and a 8.3B-parameter CogView-large. They are also general for all Transformer pretraining, and will enable the training of very deep Transformers in the future. As an evidence, we used PB-relax successfully eliminating the overflow in training  
\ifblind
\else
a 10B-parameter GLM~\cite{du2021all}.
\fi
However, in general, the precision problems in language pretraining is not so significant as in text-to-image pretraining. We hypothesize that the root is the heterogeneity of data, because we observed that text and image tokens are distinguished by scale in some hidden states. 
Another possible reason is hard-to-find underflow, guessed by DALL-E.
A thorough investigation is left for future work.

\section{Finetuning}
CogView steps further than DALL-E on finetuning. Especially, we can improve the text-to-image generation via finetuning CogView for super-resolution and self-reranking. All the finetuning tasks can be completed within one day on a single DGX-2.  

\subsection{Super-resolution}\label{sec:sr}
\begin{figure*}[h] 
    \includegraphics[width=\textwidth]{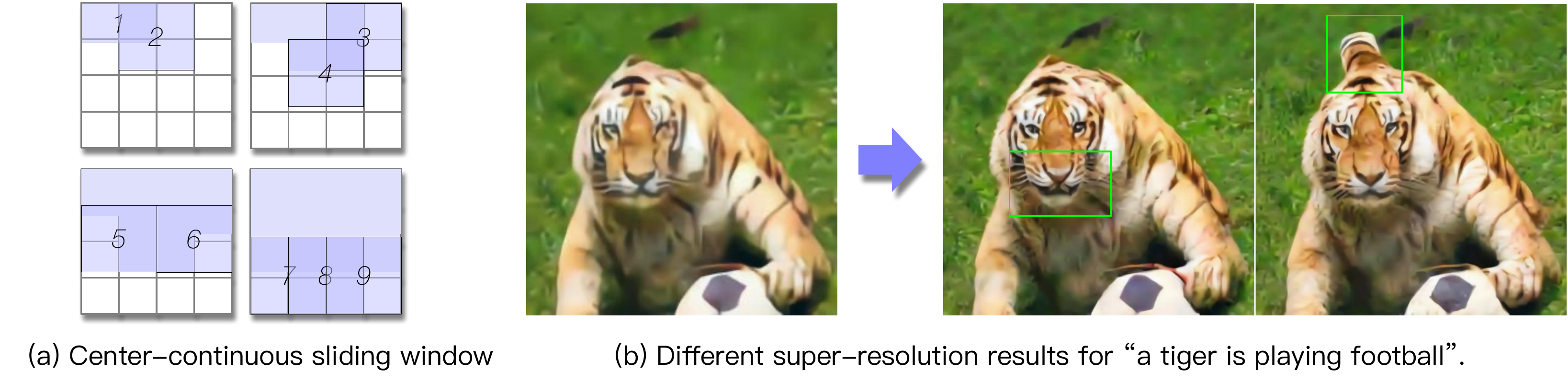}
    \caption{(a) A $64\times 64$-token image are generated patch-by-patch in the numerical order. The overlapping positions will \emph{not} be overwritten. The key idea is to make the tokens in the 2nd and 4th regions -- usually regions of faces or other important parts -- generated when attending to the whole region. (b) The finetuned super-resolution model does not barely transform the textures, but generates new local structures, e.g. the open mouth or tail in the example.}\label{fig:sr}
    \vspace{-4mm}%
\end{figure*}

Since the image tokenizer compresses $256\times256$-pixel images into $32\times32$-token sequences before training, the generated images are blurrier than real images due to the lossy compression. However, enlarging the sequence length will consume much more computation and memory due to the $O(n^2)$ complex of attention operations. Previous works~\cite{dong2014learning} about super-resolution, or image restoration, usually deal with images already in high resolution, mapping the blurred local textures to clear ones. They cannot  be applied to our case, where we need to add meaningful details to the generated low-resolution images. Figure~\ref{fig:sr} (b) is an example of our finetuning method, and illustrates our desired behavior of super-resolution.

The motivation of our finetuning solution for super-resolution is a belief that \emph{CogView is trained on the most complex distribution in general domain, and the objects of different resolution has already been covered}.\footnote{An evidence to support the belief is that if we append ``close-up view'' at the end of the text, the model will generate details of a part of the object.} Therefore, finetuning CogView for super-resolution should not be hard.

Specifically, we first finetune CogView into a conditional super-resolution model from $16\times 16$ image tokens to $32\times 32$ tokens. Then we magnify an image of $32\times 32$ tokens to $64\times 64$ tokens ($512\times 512$ pixels) patch-by-patch via a center-continuous sliding-window strategy in Figure~\ref{fig:sr} (a). This order performs better that the raster-scan order in preserving the completeness of the central area.

To prepare data, we crop about 2 million images to $256\times 256$ regions and downsample them to $128\times 128$. After tokenization, we get $32\times 32$ and $16\times 16$ sequence pairs for different resolution. The pattern of finetuning sequence is ``\texttt{[ROI1]} text tokens \texttt{[BASE]}\texttt{[BOI1]} $16\times 16$ image tokens \texttt{[EOI1]} \texttt{[ROI2]}\texttt{[BASE]} \texttt{[BOI2]} $32\times 32$ image tokens \texttt{[EOI2]}'', longer than the max position embedding index 1087. As a solution, we recount the position index from 0 at \texttt{[ROI2]}.\footnote{One might worry about that the reuse of position indices could cause confusions, but in practice, the model can distinguish the two images well, probably based on whether they can attend to a \texttt{[ROI2]} in front.} 
\subsection{Image Captioning and Self-reranking}
To finetune CogView for image captioning is straightforward: exchanging the order of text and image tokens in the input sequences. Since the model has already learnt the corresponding relationships between text and images, reversing the generation is not hard. We did not evaluate the performance due to that (1) there is no authoritative Chinese image captioning benchmark (2) image captioning is not the focus of this work. The main purpose of finetuning such a model is for self-reranking. 

We propose the \emph{Caption Loss} (CapLoss) to evaluate the correspondence between images and text. More specifically, $\text{CapLoss}(x,t)=\frac{1}{|t|}{\sum_{i=0}^{|t|}-\log p(t_i|x, t_{0:i-1})} $, where $t$ is a sequence of text tokens and $x$ is the image. $\text{CapLoss}(x,t)$ is the cross-entropy loss for the text tokens, and this method can be seen as an adaptation of inverse prompting~\cite{zou2021controllable} for text-to-image generation. Finally, images with the lowest CapLosses are chosen.

\begin{figure*}[h] 
    \includegraphics[width=\textwidth]{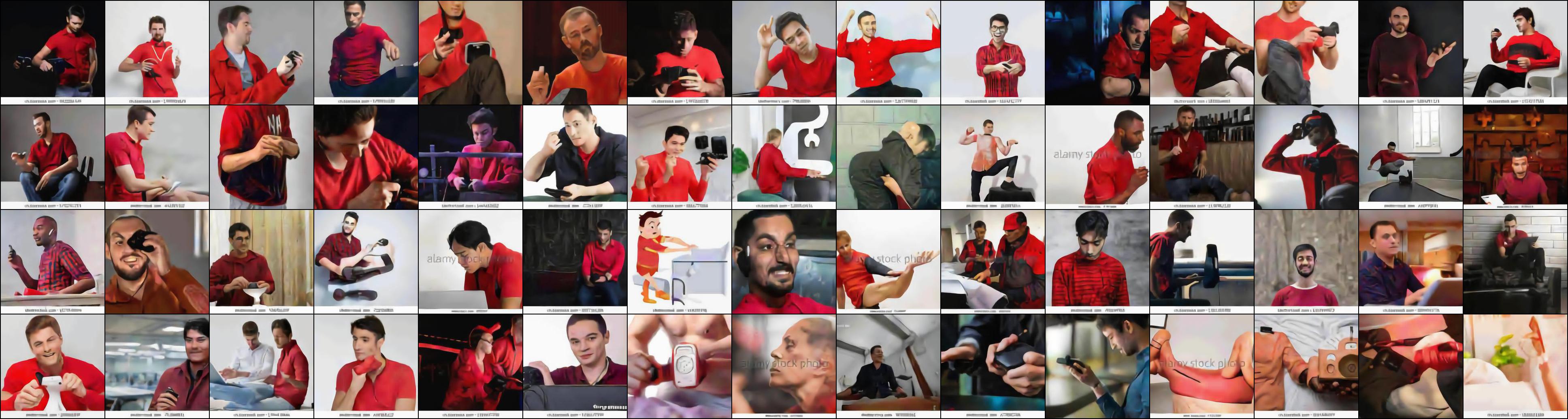}
    \caption{60 generated images for ``A man in red shirt is playing video games'' (selected at random from COCO), displayed in the order of CapLoss. Most bad cases are ranked in last places. The diversity also eases the concern that CogView might be overfitting a similar image in the training set.}\label{fig:rerank}
    \vspace{-4mm}%
\end{figure*}

Compared to additionally training another constrastive self-supervised model, e.g. CLIP~\cite{radford2021learning}, for reranking, our method consumes less computational resource because we only need finetuning. The results in Figure~\ref{fig:clip_compare} shows the images selected by our methods performs better in FID than those selected by CLIP. Figure~\ref{fig:rerank} shows an example for reranking.

\subsection{Style Learning}
Although CogView is pretrained to cover diverse images as possible, the desire to generate images of a specific style or topic cannot be satisfied well. We finetune models on four styles: Chinese traditional drawing, oil painting, sketch, and cartoon. Images of these styles are automatically extracted from search engine pages including Google, Baidu and Bing, etc., with keyword as ``An image of \texttt{\{style\} style}'', where \texttt{\{style\}} is the name of style. We finetune the model for different styles separately, with 1,000 images each.

\begin{figure*}[h]
\includegraphics[width=\textwidth]{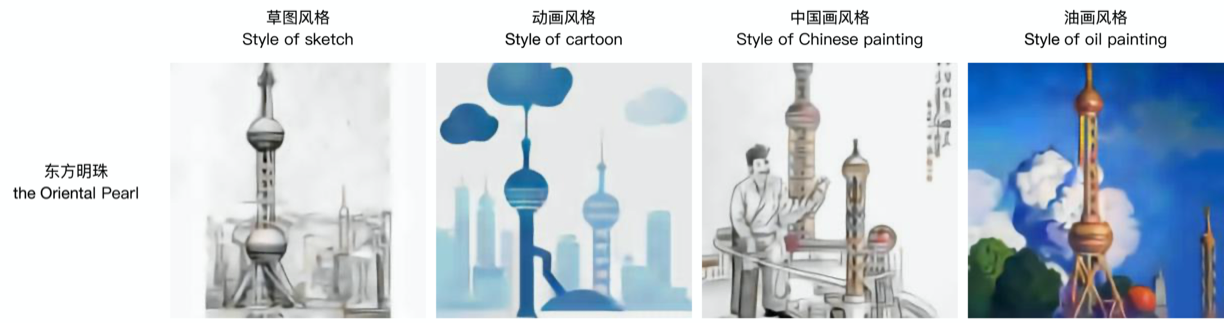}
\caption{Generated images for ``The Oriental Pearl'' (a landmark of Shanghai) in different styles.}\label{fig:style}
\vspace{-3mm}%
\end{figure*}

During finetuning, the corresponding text for the images are also ``An image of \texttt{\{style\}} style``. When generating, the text is ``A \texttt{\{object\}} of \texttt{\{style\}} style``, where $\texttt{\{object\}}$ is the object to generate. In this way, CogView can transfer the knowledge of shape of the objects learned from pretraining to the style of finetuning. Figure~\ref{fig:style} shows examples for the styles.

\subsection{Industrial Fashion Design}
\begin{wrapfigure}[13]{r}{0.47\textwidth}
    \centering
    \vspace{-6mm}%
    \includegraphics[width=0.47\textwidth]{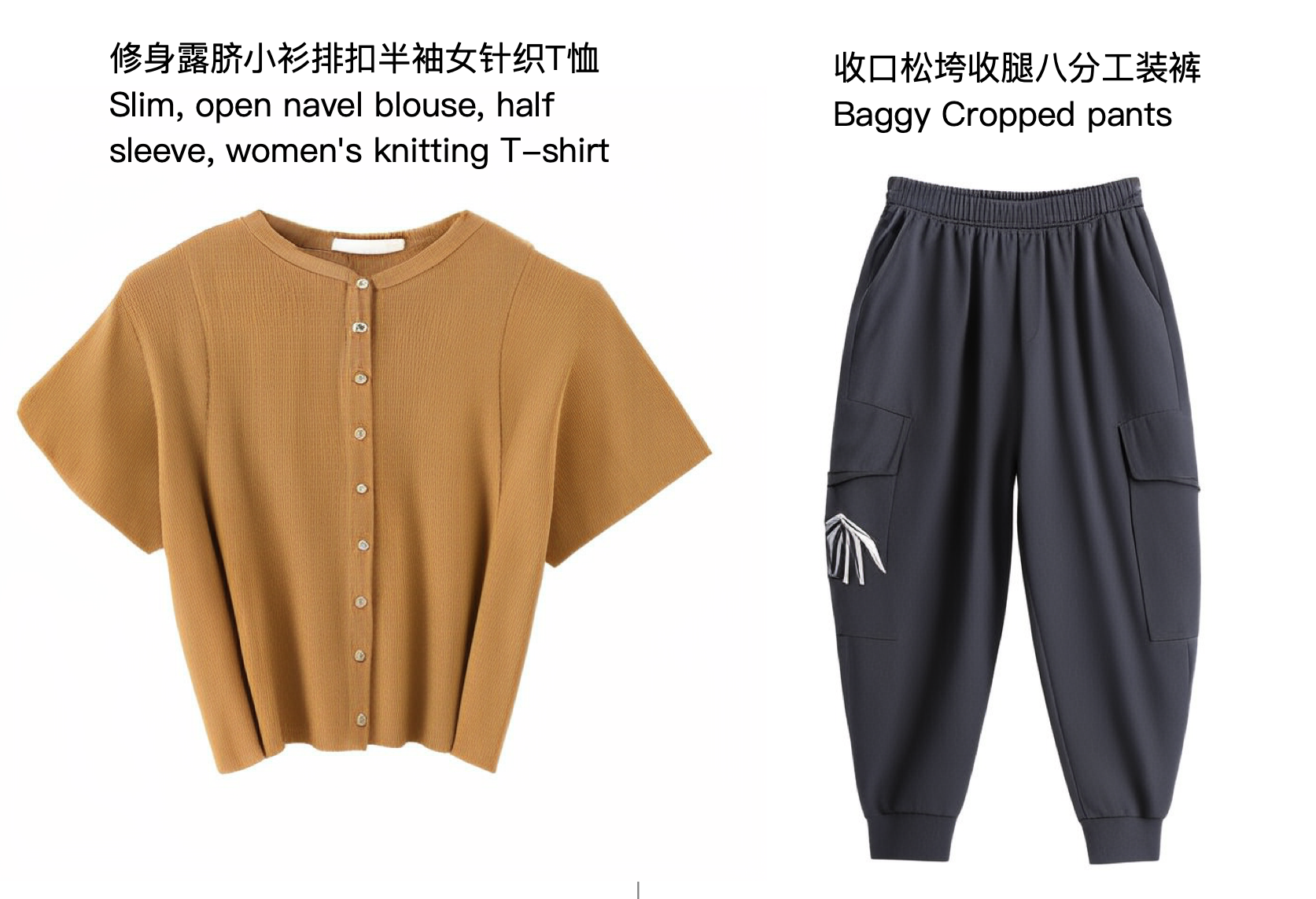} 
    \caption{Generated images for fashion design.}
    \label{fig:fs}
\end{wrapfigure}
When the generation targets at a single domain, the complexity of the textures are largely reduced. In these scenarios, we can (1) train a VQGAN~\cite{esser2020taming} instead of VQVAE for the latent variable for more realistic textures, (2) decrease the number of parameters and increase the length of sequences for a higher resolution. Our \emph{three-region sparse attention}
\ifblind
\else
(Appendix~\ref{app:sparse})
\fi
 can speed up the generation of high-resolution images in this case.

We train a 3B-parameter model on about 10 million fashion-caption pairs, using $50\times 50$ VQGAN image tokens and decodes them into $800\times 800$ pixels. Figure~\ref{fig:fs} shows samples of CogView for fashion design, which has been successfully deployed to 
\ifblind
\else 
Alibaba Rhino fashion production. 
\fi

\vspace{-2mm}
\section{Experimental Results}
\vspace{-1mm}
\subsection{Machine Evaluation}
\vspace{-1mm}
At present, the most authoritative machine evaluation metrics for general-domain text-to-image generation is the FID on MS COCO, which is not included in our training set. To compare with DALL-E, we follow the same setting, evaluating CogView on a subset of 30,000 captions sampled from the dataset, after applying a Gaussian filter with varying radius to both the ground-truth and generated images.\footnote{We use the same evaluation codes with DM-GAN and DALL-E, which is available at \url{https://github.com/MinfengZhu/DM-GAN}.} The captions are translated into Chinese for CogView by machine translation. To fairly compare with DALL-E, we do not use super-resolution. Besides, DALL-E generates 512 images for each caption and selects the best one by CLIP, which needs to generate about 15 billion tokens. To save computational resource, we select the best one from 60 generated images according to their CapLosses. The evaluation of CapLoss is on a subset of 5,000 images. We finally enhance the contrast of generated images by 1.5. Table~\ref{tab:fid} shows the metrics for CogView and other methods.
\begin{table}[h]
    \centering
    \vspace{-2mm}%
    \caption{Metrics for machine evaluation. Statistics about DALL-E and GANs are extracted from their figures. FID-$k$ means that all the images are blurred by a Gaussian Filter with radius $k$.}\label{tab:fid}
    \begin{tabular}[]{cccccccc}
    \toprule 
        Model & FID-0 &  FID-1 &  FID-2 &  FID-4 &  FID-8 & IS & CapLoss \\ \midrule
		AttnGAN & 35.2  & 44.0 & 72.0 & 108.0 & 100.0 & 23.3& 3.01\\ 
		DM-GAN & \textbf{26.5}  & 39.0 & 73.0 & 119.0 & 112.3 & \textbf{32.2} & 2.87 \\
		DF-GAN & \textbf{26.5}  &  33.8 & 55.9 & 91.0 & 97.0 & 18.7 & 3.09\\
		DALL-E & 27.5 & 28.0 & 45.5 & 83.5 & 85.0 & 17.9 & --- \\
        \midrule
		CogView & 27.1 & \textbf{19.4} & \textbf{13.9} & \textbf{19.4} & \textbf{23.6} & 18.2 & \textbf{2.43}\\  \bottomrule
    \end{tabular}
    \vspace{-3mm}%
\end{table} 


\textbf{Caption Loss as a Metric.} FID and IS are designed to measure the quality of unconditional generation from relatively simple distributions, usually single objects. However, text-to-image generation should be evaluated pair-by-pair. Table~\ref{tab:fid} shows that DM-GAN achieves the best unblurred FID and IS, but is ranked last in human preference (Figure~\ref{fig:stat}(a)). Caption Loss is an absolute (instead of relative, like CLIP) score, so that it can be averaged across samples. It should be a better metrics for this task and is more consistent with the overall scores of our human evaluation in \S~\ref{sec:human}. 
\begin{wrapfigure}[8]{r}{0.5\textwidth}
    \centering
    \vspace{-3mm}%
    \includegraphics[width=0.5\textwidth]{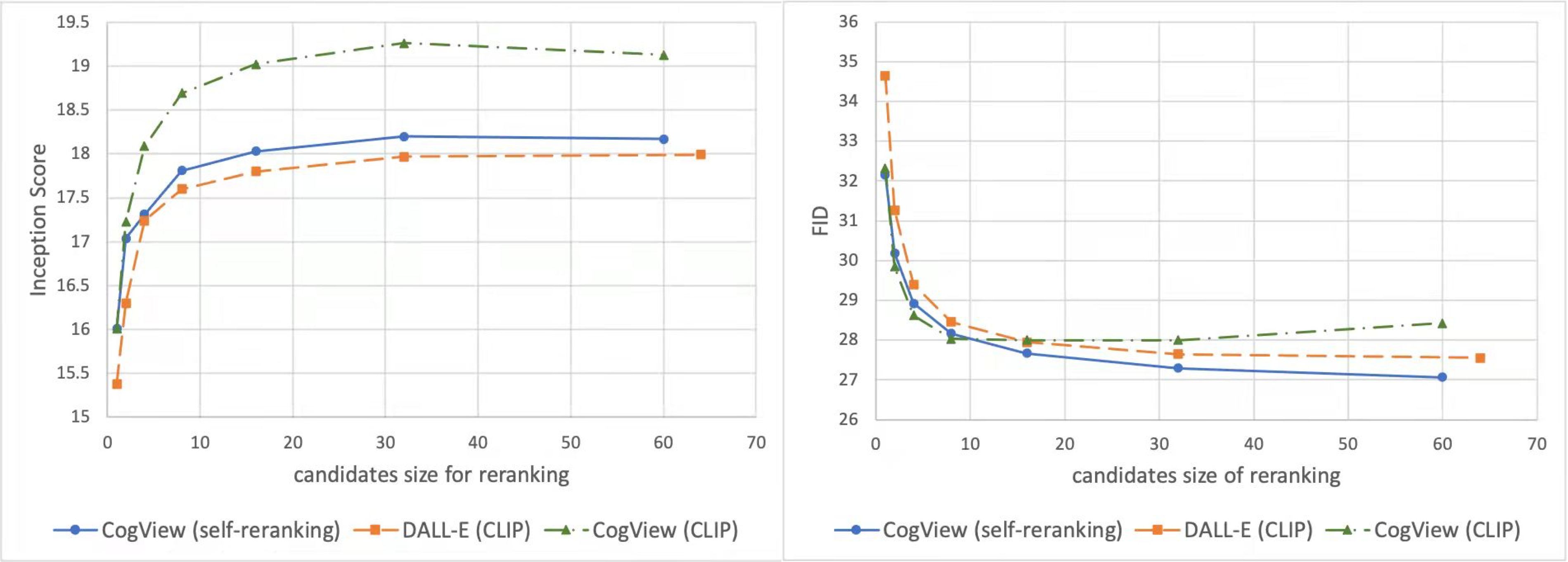} 
    \vspace{-5mm}%
    \caption{IS and FID-0 for CLIP and self-ranking. }
    \label{fig:clip_compare}
  \end{wrapfigure}
\textbf{Comparing self-reranking with CLIP.} We evaluate the FID-0 and IS of CogView-generated images selected by CLIP and self-reranking on MS COCO. Figure~\ref{fig:clip_compare} shows the curves with different number of candidates. Self-reranking gets better FID, and steadily refines FID as the number of candidates increases. CLIP performs better in increasing IS, but as discussed above, it is not a suitable metric for this task.

\textbf{Discussion about the differences in performance between CogView and DALL-E.} Since DALL-E is pretrained with more data and parameters than CogView, why CogView gets a better FID even without super-resolution? It is hard to know the accurate reason, because DALL-E is not open-source, but we guess that the reasons include: (1) CogView uses PB-relax and Sandwich-LN for a more stable optimization. (2) DALL-E uses many cartoon and rendered data, making the texture of generated images quite different from that of the photos in MS COCO. (3) Self-reranking selects images better in FID than CLIP. (4) CogView is trained longer  (96B trained tokens in CogView vs. 56B trained tokens in DALL-E).
\subsection{Human Evaluation} \label{sec:human}
Human evaluation is much more persuasive than machine evaluation on text-to-image generation. Our human evaluation consists of 2,950 groups of comparison between images generated by AttnGAN, DM-GAN, DF-GAN, CogView, and recovered ground truth, i.e., the ground truth blurred by our image tokenizer. Details and example-based comparison between models are in 
\ifblind
\else 
Appendix~\ref{app:human}. 
\fi

\begin{figure*}[h] 
    \includegraphics[width=\textwidth]{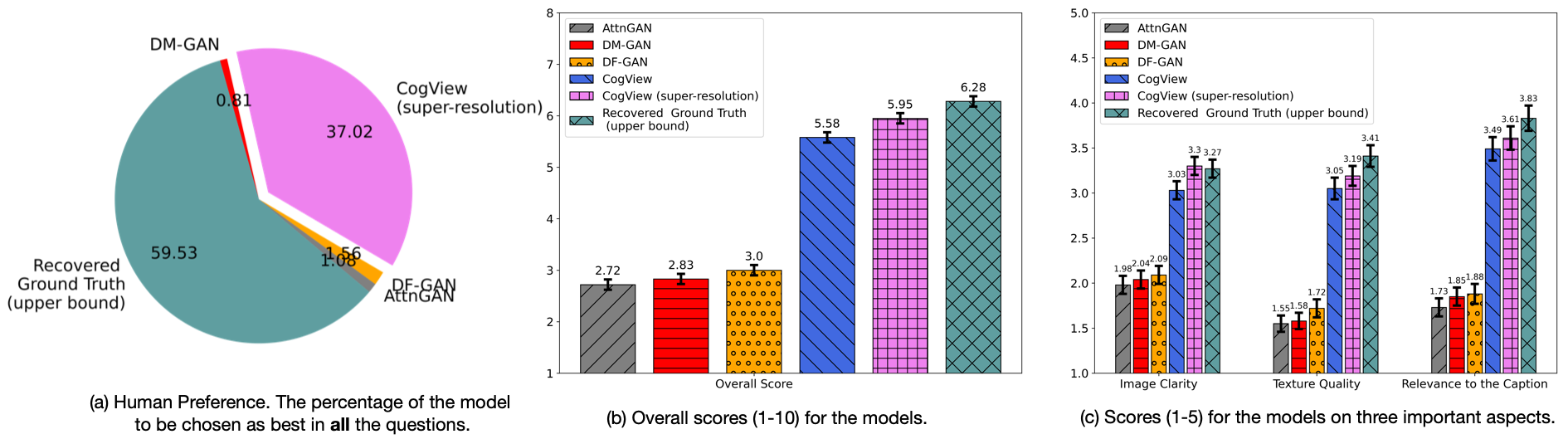}
    \caption{Human Evaluation results. The recovered ground truth is obtained by first encoding the ground truth image and then decoding it, which is theoretically the upper bound of CogView. }\label{fig:stat}
    \vspace{-4mm}%
\end{figure*}

Results in Figure~\ref{fig:stat} show that CogView outperforms GAN-based baselines at a large margin. CogView is chosen as the best one with probability 37.02\%, competitive with the performance of recovered ground truth (59.53\%). Figure~\ref{fig:stat}(b)(c) also indicates our super-resolution model consistently improves the quality of images, especially the clarity, which even outperforms the recovered ground truth.
\section{Conclusion and Discussion}
\textbf{Limitations.} A disadvantage of CogView is the slow generation, which is common for auto-regressive model, because each image is generated token-by-token. The blurriness brought by VQVAE is also an important limitation. These problems will be solved in the future work.

\textbf{Ethics Concerns.} Similar to Deepfake, CogView is vulnerable to malicious use~\cite{westerlund2019emergence} because of its controllable and strong capacity to generate images. The possible methods to mitigate this issue are discussed in a survey~\cite{brundage2018malicious}. Moreover, there are usually fairness problems in generative models about human~\footnote{\url{https://thegradient.pub/pulse-lessons}}. In Appendix~\ref{app:fairness}, we analyze the situation about fairness in CogView and introduce a simple ``word replacing'' method to solve this problem.

We systematically investigate the framework of combining VQVAE and Transformers for text-to-image generation. CogView demonstrates promising results for scalable cross-modal generative pretraining, and also reveals and solves the precision problems probably originating from data heterogeneity. We also introduce methods to finetune CogView for diverse downstream tasks. We hope that CogView could advance both research and application of controllable image generation and cross-modal knowledge understanding, but need to prevent it from being used to create images for misinformation.

\acksection
We would like to thank Zhao Xue, Zhengxiao Du, Hanxiao Qu, Hanyu Zhao, Sha Yuan, Yukuo Cen, Xiao Liu, An Yang, Yiming Ju for their help in data, machine maintaining or discussion. We would also thank Zhilin Yang for presenting this work at the conference of BAAI. 

Funding in direct support of this work: a fund for GPUs donated by BAAI, a research fund from Alibaba Group, NSFC for Distinguished Young Scholar 
(61825602),
NSFC (61836013).

\bibliography{ref} 
\bibliographystyle{abbrvnat}
\ifblind
\else
\newpage
\appendix

\section{Data Collection and Details about the Tokenizers}\label{app:data}
We collected about 30 million text-image pairs from multiple channels, and built a 2.5TB new dataset (after tokenization, the size becomes about 250GB). The dataset is an extension of 
\ifblind
an open-dataset project\footnote{\texttt{[hidden during review]}}
\else
project WudaoCorpora~\cite{WudaoCorpora}\footnote{\url{https://wudaoai.cn/data}}.
\fi
About 50\% of the text is in English, including Conceptual Captions~\cite{sharma2018conceptual}. They are translated into Chinese by machine translation. In addition, we did not remove the watermarks and white edges in the dataset even though they affect the quality of generated images, because we think it will not influence the conclusions of our paper from the perspective of research.

The sources of data are basically classified into the following categories: (1) Professional image websites (both English and Chinese). The images in the websites are usually with captions. Data from this channel constitute the highest proportion. (2) Conceptual Captions~\cite{sharma2018conceptual} and ImageNet~\cite{deng2009imagenet}. (3) News pictures online with their surrounding text. (4) A small part of item-caption pairs from 
\ifblind
a large e-commerce company
\else
Alibaba
\fi
. (5) Image search engines. In order to cover as many common entities as possible, we made a query list consist of 1,200 queries. Every query was an entity name extracted from a large-scale knowledge graph. We choose seven major categories: food, regions, species, people names, scenic, products and artistic works. We extracted top-$k$ entities for each category based on their number of occurrences in the English Wikipedia, where $k$ is manually selected for each category. We collected the top-100 images returned by every major search engine website for each query. 

We have already introduced tokenizers in section~\ref{sec:tokenizer}, and here are some details. The text tokenizer are directly based on the SentencePiece package at \url{https://github.com/google/sentencepiece}. The encoder in the image tokenizer is a 4-layer convolutional neural network (CNN) with 512 hidden units and ReLU activation each layer. The first three layers have a receptive field of 4 and stride of 2 to half the width and height of images, and the final layer is a $1\times 1$ convolution to transform the number of channels to 256, which is the hidden size of embeddings in the dictionary. The decoder have the same architecture with the encoder except replacing convolution as deconvolution. The embeddings in the dictionary are initialized via Xavier uniform initialization~\cite{glorot2010understanding}.
\section{Sparse Attention}\label{app:sparse}
\begin{wrapfigure}[25]{r}{0.43\textwidth}
    \centering
    \vspace{-8mm}%
    \includegraphics[width=0.43\textwidth]{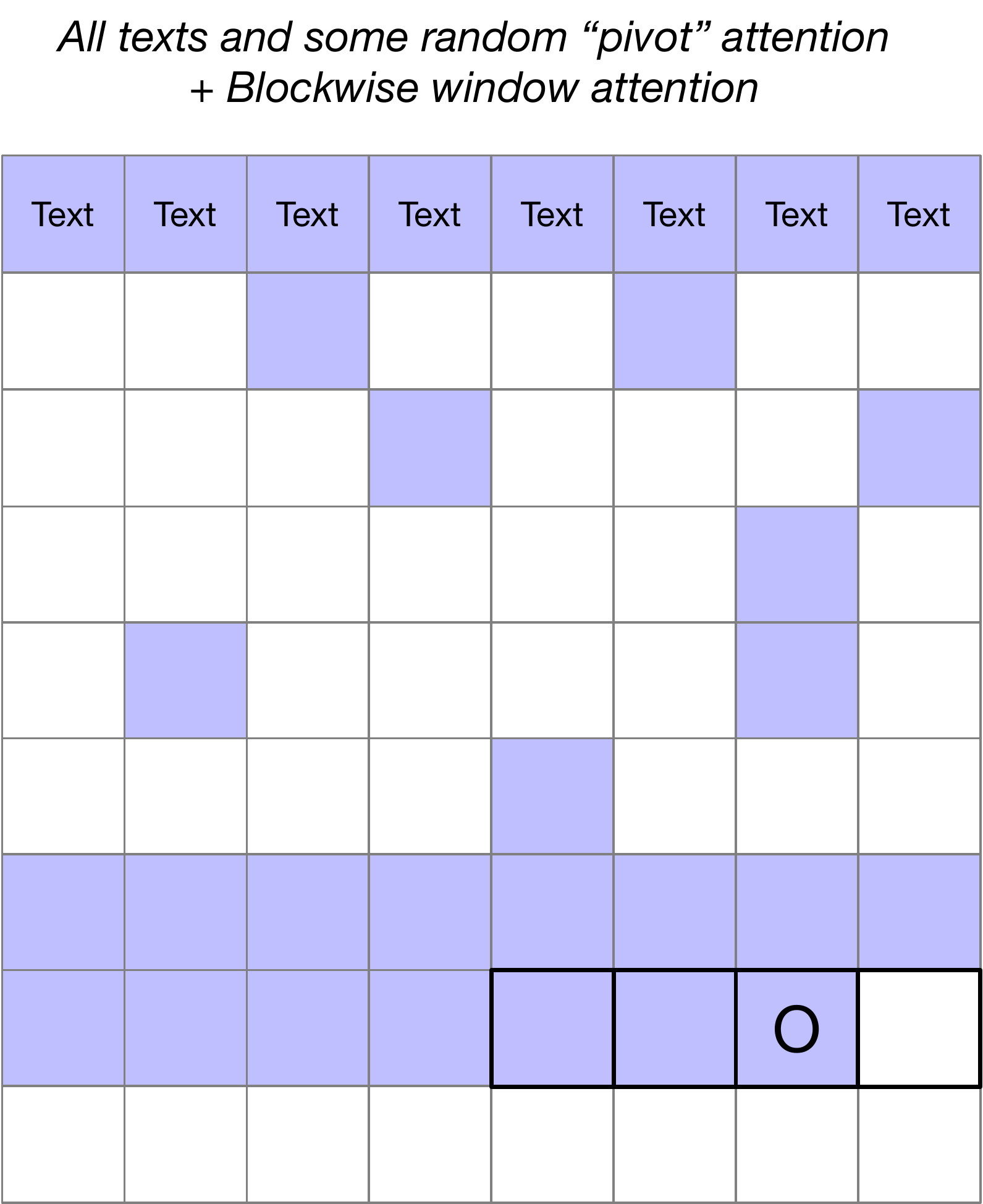} 
    \caption{Illustration about our three-region sparse attention. The sequence is shown as a $H\times W$ image and some text tokens in front. Colored grids are all the tokens attended to by the token marked ``O''. In this case, each block consists of four consecutive tokens. }\label{fig:sparse}
\end{wrapfigure}
As shown in Figure~\ref{fig:sparse}, we design the \emph{three-region sparse attention}, an implementation-friendly sparse attention for text-to-image generation. Each token attends to all text tokens, all \emph{pivots} tokens and tokens in the blocks in an adjacent window before it. 

The pivot tokens are image tokens selected at random, similar to big bird~\cite{zaheer2020big}. They are re-sampled every time we enter a new layer. We think they can provide global information about the image. 

The blockwise window attention provides local information, which is the most important region. The forward computation of 1-D window attention can be efficiently implemented inplace by carefully padding and altering the strides of tensors, because the positions to be attended are already continguous in memory. However, we still need extra memory for backward computation if without customized CUDA kernels. We alleviate this problem by grouping adjacent tokens into blocks, in which all the tokens attend to the same tokens (before causally masking). More details are included in our released codes.

In our benchmarking on sequences of 4096 tokens, the three-region sparse attention (768 text and pivot tokens, 768 blockwise window tokens) is $2.5\times$ faster than vanilla attention, and saves $40\%$ GPU memory. The whole training is $1.5\times$ faster than that with vanilla attention and saves $20\%$ GPU memory. With the same hyperparameters, data and random seeds, their loss curves are nearly identical, which means the sparse attention will not influence the convergence.  

However, we did not use three-region sparse attention during training the 4-billion-parameter CogView, due to the concern that it was probably not compatible with finetuning for super-resolution in section~\ref{sec:sr}. But it successfully accelerated the training of CogView-fashion without side effects.


\section{Attention Analysis} \label{app:attn}
To explore the attention mechanism of CogView, we visualize the attention distribution during inference by plotting heat maps and marking the most attended tokens. We discover that our model's attention heads exhibit strong ability on capturing both position and semantic information, and attention distribution varies among different layers. The analysis about the scale of attention scores is in section~\ref{app:attnscale}.

\subsection{Positional Bias}

The attention distribution is highly related to images' position structures. There are a lot of heads heavily attending to fixed positional offsets, especially multiple of 32 (which is the number of tokens a row contains) (Figure~\ref{fig:attention-heatmap} (a)). Some heads are specialized to attending to the first few rows in the image (Figure\ref{fig:attention-heatmap} (b)) . Some heads' heat maps show checkers pattern (Figure~\ref{fig:attention-heatmap} (c)), indicating tokens at the boundary attends differently from that at the center. Deeper layers also show some broad structural bias. For example, some heads attend heavily on tokens at top/lower half or the center of images (Figure~\ref{fig:attention-heatmap} (d)(e)). 

\begin{figure*}[h] 
	\includegraphics[width=\textwidth]{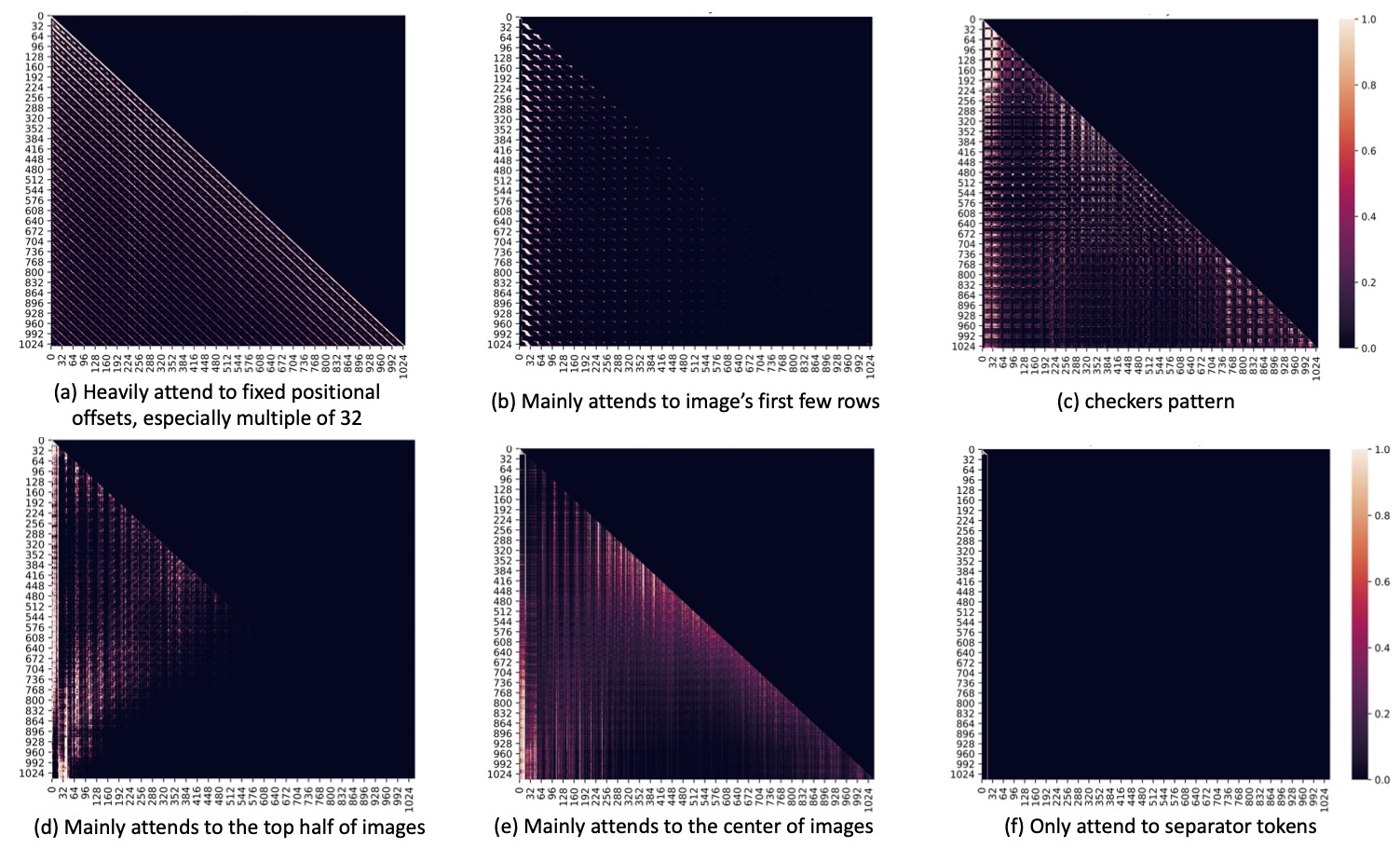}
	\caption{(a)(b)(c) Our model's attention is highly related to images' positional structures. (d)(e) Our model's attention show some broad structural bias. (f) Some heads only attend to a few tokens such as separator token.}\label{fig:attention-heatmap}
\end{figure*}

\subsection{Semantic Segmentation}

\begin{figure*}[h] 
\centering

	\includegraphics[width=0.6\textwidth]{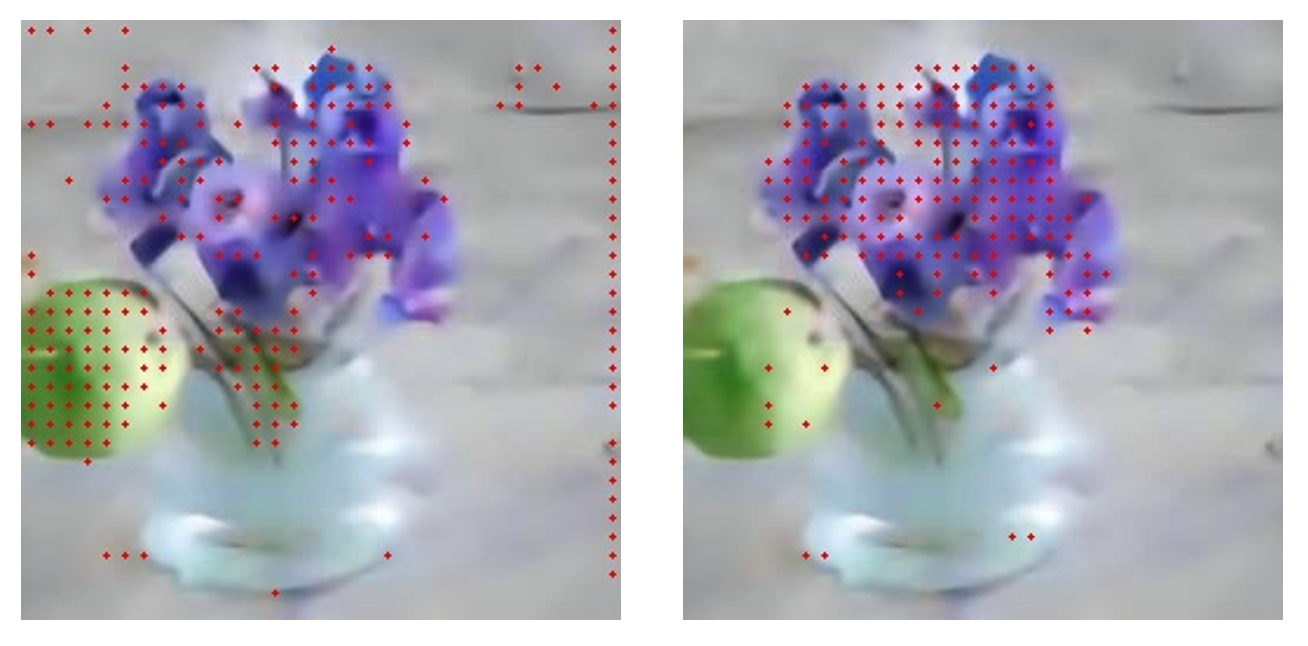}
	\caption{Our model's attention heads successfully captured items like apple and purple flowers. Pixels corresponding to the most highly attended tokens are marked with red dots.}\label{fig:attention-hotspot}
\end{figure*}

	

The attention in CogView also shows that it also performs implicit semantic segmentation. Some heads highlight major items mentioned in the text. We use "There is an apple on the table, and there is a vase beside it, with purple flowers in it." as input of our experiment. In Figure~\ref{fig:attention-hotspot} we marked pixels corresponding to the most highly attended tokens with red dots, and find that attention heads successfully captured items like apple and purple flowers.

\subsection{Attention Varies with Depth}
Attention patterns varies among different layers. Earlier layers focus mostly on positional information, while later ones focus more on the content. Interestingly, we observe that attention become sparse in the last few layers (after layer 42), with a lot of heads only attend to a few tokens such as separator tokens (Figure~\ref{fig:attention-heatmap} (f)). One possible explanation is that those last layers tend to concentrate on current token to determine the output token, and attention to separator tokens may be used as a no-op for attention heads which does not substantially change model's output, similar to the analysis in BERT~\cite{clark2019does}.  As the result, the last layers' heads disregard most tokens and make the attention layers degenerate into feed-forward layers. 

\subsection{Value Scales of Attention}\label{app:attnscale}
As a supplement to section~\ref{sec:stable}, we visualize the value scales of attention in the 38-th layer, which has the largest scale of attention scores $Q^TK/\sqrt{d}$ in CogView. The scales varies dramatically in different heads, but the variance in each single head is small (that is why the attention does not degenerate, even though the scores are large). We think the cause is that the model wants different \emph{sensitiveness} in different heads, so that it learns to multiply different constants to get $Q$ and $K$. As a side effect, the values may have a large bias. The PB-relax for attention is to remove the bias during computation. 
\begin{figure*}[h]
    \centering
\includegraphics[width=\textwidth]{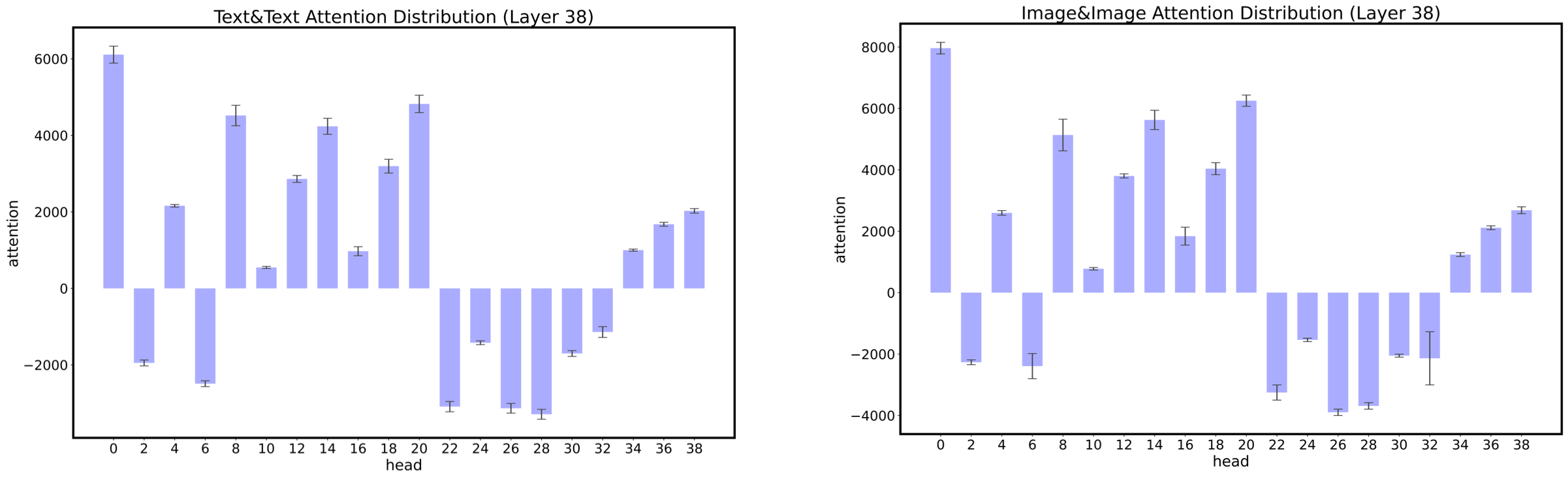}
\caption{Illustration of scales of attention scores in the 38-th layer. Only half are heads are shown for display reasons. The error bar is from the minimum to the maximum of scores. The values of text-to-text attention scores are smaller, indicating the scales are related to the data.}\label{fig:scale}
\end{figure*}

\section{Fairness in CogView: Situation and Solution}\label{app:fairness}
\begin{figure*}[h]
    \centering
\includegraphics[width=0.8\textwidth]{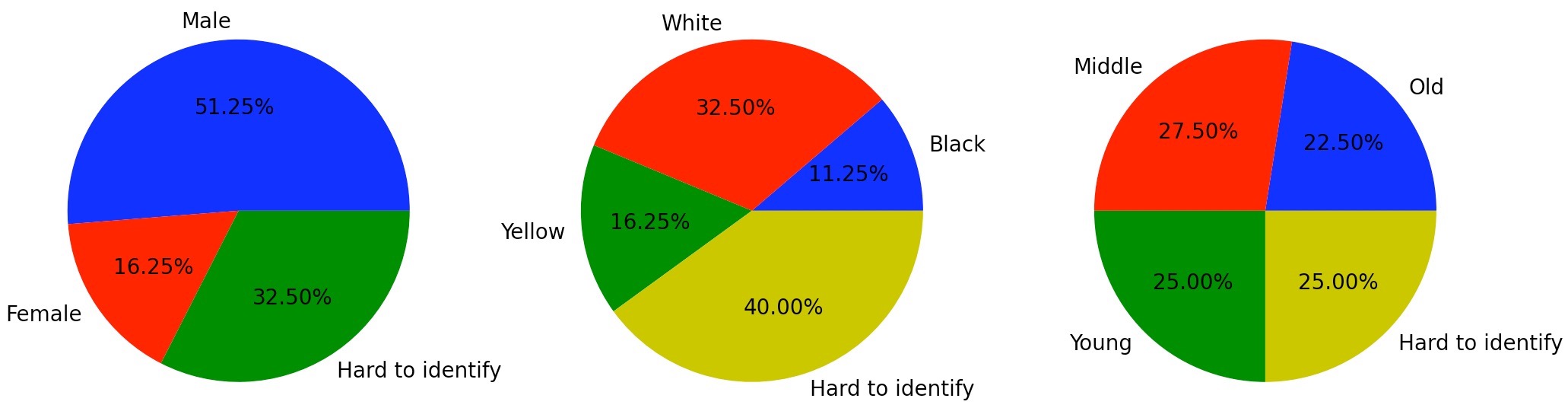}
\caption{The distribution of different genders, races and ages of the generation of ``a face, photo''.}\label{fig:fairness}
\end{figure*}
\textbf{Evaluation of the situation of fairness in CogView.} We examine the bias in the proportion of different races and genders. Firstly, if given the detailed description in the text, e.g. a black man or an Asian woman, CogView can generate correctly for almost all samples.
We also measure the proportion of the generated samples without specific description by the text ``a face, photo''. The figure of proportions in different races and genders are in Figure~\ref{fig:fairness}. The (unconditional) generated faces are relatively balanced in races and ages, but with more men than women due to the data distribution.

CogView is also beset by the bias in gender due to the stereotypes if not specifying the gender. However if we specify the gender, almost all the gender and occupation are correct. We tested the examples introduced in~\cite{caliskan2017semantics}, and generated images for the text \{male, female\} $\times$ \{``science'', mathematics'', ``arts'', ``literature''\}. Results are showed in \href{https://i.imgur.com/EuOmJfU.jpg}{this outer link} to reduce the size of our paper.

\textbf{Word Replacing Solution.} Different from the previous unconditional generative models, we have a very simple and effective solution for racial and gender fairness. 

We can directly add some adjective words sampled from ``white'', ``black'',  ``Asian'', ..., and ``male'', ``female'' (if not specified) in the front of the words for human, like ``people'' or ``person'', in the text. The sampling is according to the real proportion in the whole population. We can train an additional NER model to find the words about human.

Since CogView will predict correctly according to the results above, if given description, this method will greatly help solve the fairness problem in generative models.
\section{Details about Human Evaluation}
\label{app:human}
To evaluate the performance, we conduct a human evaluation to make comparisons between various methods, similar to previous works~\cite{koh2021text,ramesh2021zero}. In our designed evaluation, 50 images and their captions are randomly selected from the MS COCO dataset. For each image, we use the caption to generate images based on multiple models including AttnGAN, DM-GAN, DF-GAN and CogView. We do not generate images with DALL-E as their model has not been released yet. For each caption, evaluators are asked to give scores to 4 generated images and the recovered ground truth image respectively. The recovered ground truth image refers to the image obtained by first encoding the ground truth image (the original image in the MS COCO dataset after cropped into the target size) and then decoding it.

For each image, evaluators first need to give 3 scores ($1\sim 5$) to evaluate the image quality from three aspects: the image clarity, the texture quality and the relevance to the caption. Then, evaluators will give an overall score ($1\sim 10$) to the image. After all 5 images with the same caption are evaluated, evaluators are required to select the best image additionally.

72 anonymous evaluators are invited in the evaluation. To ensure the validity of the evaluation results, we only collect answers from evaluators who complete all questions and over 80\% of the selected best images are accord with the one with the highest overall quality score. Finally, 59 evaluators are kept. Each evaluator is awarded with 150 yuan for the evaluation. There is no time limit for the answer.

To further evaluate the effectiveness of super-resolution, we also introduced a simple A-B test in the human evaluation. Evaluators and captions are randomly divided into two groups $E_a,E_b$ and $C_a,C_b$ respectively. For evaluators in $E_a$, the CogView images with captions from $C_a$ are generated without super-resolution while those from $C_b$ are generated with super-resolution. The evaluators in $E_b$ do the reverse. Finally, we collected equal number of evaluation results for CogView images with and without super-resolution.

The average scores and their standard deviation are plotted in Figure~\ref{fig:stat}. Several examples of captions and images used in the human evaluation are listed in Figure~\ref{fig:human-evaluation-examples}. The evaluation website snapshots are displayed in Figure~\ref{fig:human-evaluation-website}.

\begin{figure*}[h] 
    \includegraphics[width=0.95\textwidth]{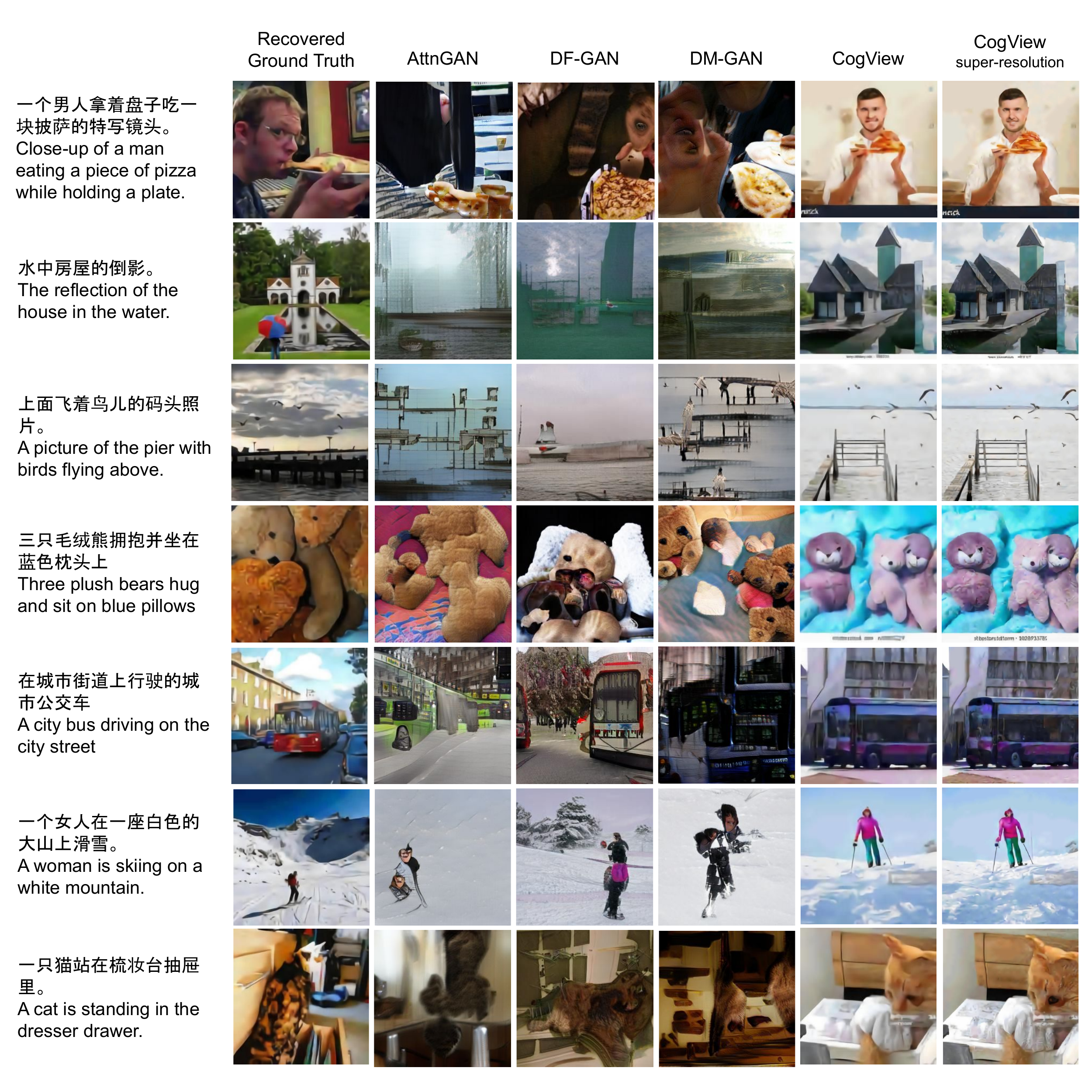}
    \caption{Human evaluation examples. The captions for evaluation are selected at random from MS COCO.}\label{fig:human-evaluation-examples}
\end{figure*}

\begin{figure*}[h] 
    \includegraphics[width=\textwidth]{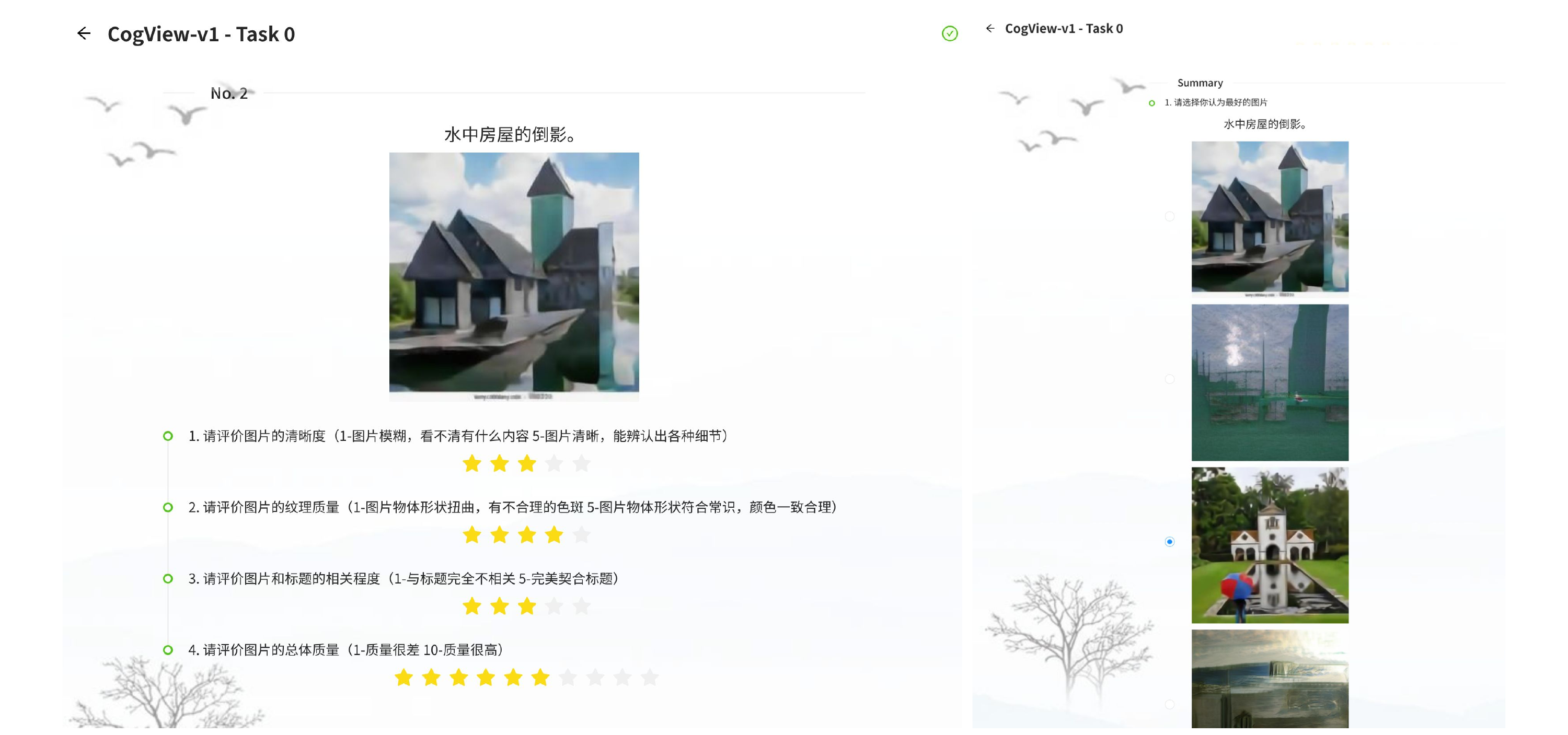}
    \caption{Snapshots of the human evaluation website. The left side is the scoring page for images and the right side is the best-selection page for all images with the same caption.}\label{fig:human-evaluation-website}
\end{figure*}

%






\section{Show Cases for captions from MS COCO}\label{app:case}
In Figure~\ref{fig:case}, we provide further examples of CogView on MS COCO. 
\begin{figure*}[h]
    \centering
\includegraphics[width=\textwidth]{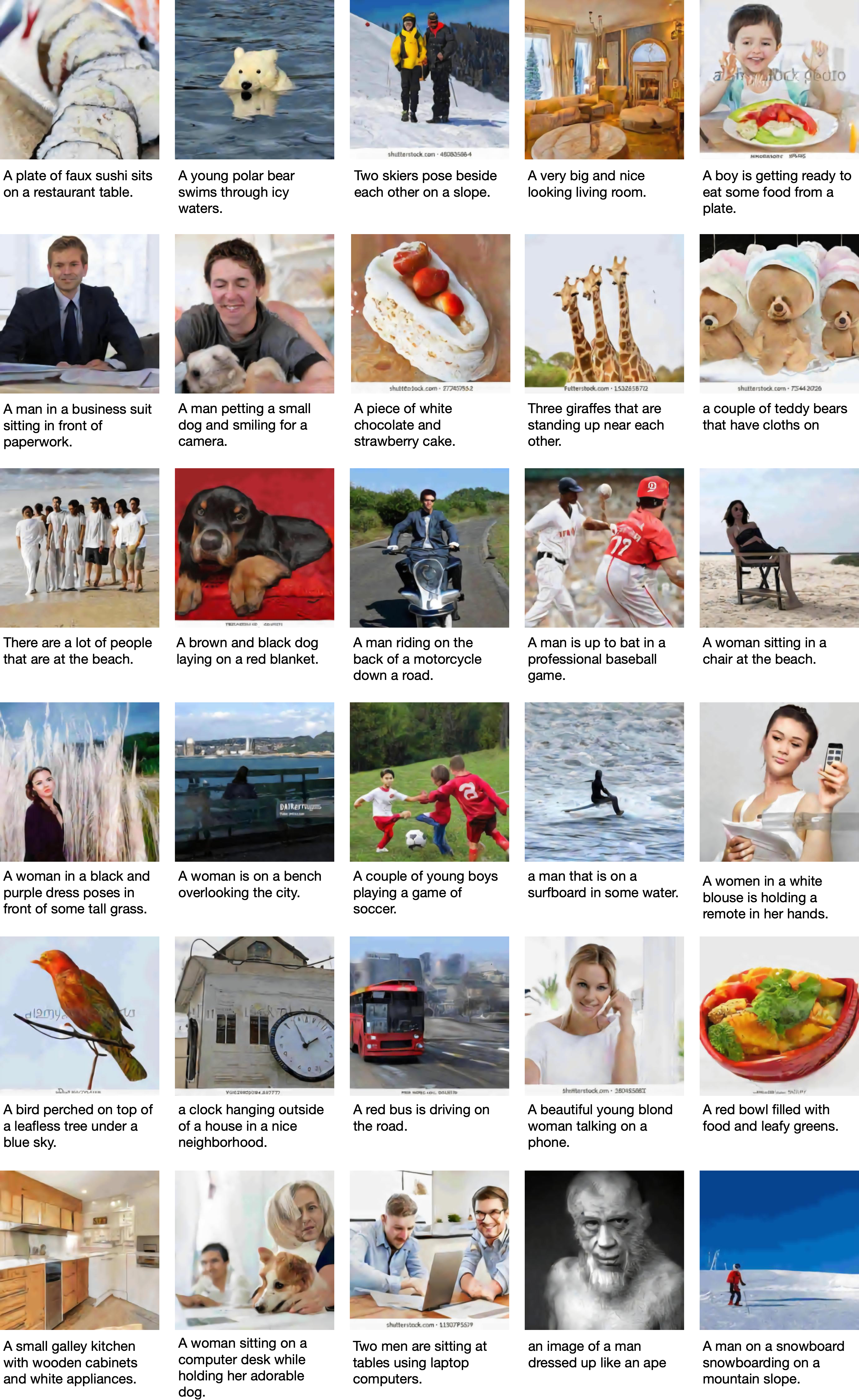}
\caption{More generated images for COCO captions (after super-resolution).}\label{fig:case}
\end{figure*}

\fi

\end{document}